\documentclass[11pt,a4paper]{article}

\PassOptionsToPackage{breaklinks}{hyperref}
\usepackage{hyperref}
\usepackage{breakurl} 

\usepackage{times}
\usepackage{latexsym}
\usepackage{amsfonts}
\usepackage{float}
\usepackage{multirow}
\usepackage{comment}
\usepackage{textcomp}
\usepackage{microtype}
\usepackage{xurl} 

\usepackage[hyperref]{acl2020}

\aclfinalcopy 

\newcommand\BibTeX{B\textsc{ib}\TeX}

\newcommand{\angles}[1]{\emph{$\langle$#1$\rangle$}}
\newcommand{\bpe}[0]{{\small @@}~~}

\title{Denoising Pre-Training and Data Augmentation Strategies for Enhanced RDF Verbalization with Transformers} 

\author{Sebastien Montella\Thanks{Both authors contributed equally.} \\
  Orange Labs, Lannion \\
  \texttt{sebastien.montella@orange.com} \\\And
  Betty Fabre\footnotemark[1] \\
  Orange Labs, Lannion \\
  \texttt{betty.fabre@orange.com} \\\AND
  Tanguy Urvoy \\
  Orange Labs, Lannion \\
  \texttt{tanguy.urvoy@orange.com} \\\And
  Johannes Heinecke \\
  Orange Labs, Lannion \\
  \texttt{johannes.heinecke@orange.com} \\\AND
  Lina Rojas-Barahona \\
  Orange Labs, Lannion \\
  \texttt{linamaria.rojasbarahona@orange.com}}

\date{}

\begin{document}
\maketitle

\begin{abstract}
The task of verbalization of RDF triples has known a growth in popularity due to the rising ubiquity of Knowledge Bases (KBs). The formalism of RDF triples is a simple and efficient way to store facts at a large scale. However, its abstract representation makes it difficult for humans to interpret. For this purpose, the WebNLG challenge aims at promoting automated RDF-to-text generation. We propose to leverage pre-trainings from augmented data with the Transformer model using a data augmentation strategy. Our experiment results show a minimum relative increases of 3.73\%, 126.05\% and 88.16\% in BLEU score for seen categories, unseen entities and unseen categories respectively over the standard training.

\end{abstract}

\section{Introduction}
The purpose of Knowledge Bases is to store unstructured data into a well-defined and highly structured representation. A commonly used formalism is the Resource Description Framework (RDF) where knowledge takes the form of triplets {$\langle$subject; predicate; object$\rangle$}. For instance if we take the triple \angles{Alan\_Bean; birthDate; 1932}, the subject is \emph{Alan\_Bean}, the predicate/relation/property is \emph{birthDate} and the object is the date \emph{1932}. 
This triple-based representation allows one to describe complex relations between entities. It is hence frequently used in natural language processing (NLP) applications such as question answering, summarization, recommendation or search engines.

Although the RDF format is an efficient method to store knowledge, triples are hard to interpret by humans. The automatic translation from sets of triples into human language is however a difficult task. Thanks to the WebNLG challenge \cite{gardent-etal-2017-webnlg}, this RDF-to-text task attracts now many researchers. This challenge aims at generating a faithful and descriptive textual content from RDF inputs. Given the triples \textit{$\langle$Alan\_Bean; birthPlace; Wheeler,\_Texas$\rangle$} and \textit{$\langle$Alan\_Bean; birthDate; 1932-03-15$\rangle$}, one possible verbalization is \textit{``Alan Bean was born in Wheeler, Texas on Mar 15, 1932.''}. Verbalization of structured data can provide an unambiguous and simple translation from abstract representations.

The WebNLG dataset is frequently used as a benchmark for Information Extraction (IE), Natural Language Understanding (NLU) or Natural Language Generation (NLG) tasks. It offers different lexicalization for the same set of triples. Despite limited in size, the high-quality of both its textual content and data-to-text mapping significantly stimulated both RDF-to-text, \textit{i.e.} NLG, and text-to-RDF, \textit{i.e.} IE and NLU, research directions. For instance, \citet{iso-etal-2020-fact} and \citet{grah-augmented-ijcai2020} have obtained state-of-the-art performances on the WebNLG RDF-to-text task. On the other hand, \citet{Guo-CycleGT-2020} considered both NLG and IE objectives within the same learning framework via cycle training. Furthermore, the exploration of a new evaluation metric is also investigated on this WebNLG dataset, confirming its attractiveness for the research community \cite{DBLP:conf/acl/DhingraFPCDC19, sellam-etal-2020-bleurt}.

Recently, the field of NLP has been significantly impacted by large pre-trained language models and novel learning architectures that improved state-of-the-art results on several benchmarks \cite{devlin-etal-2019-bert, zhang-etal-2019-ernie, zhilin-yang-xlnet-NIPS2019,radford2019language, lewis-etal-2020-bart, kevin-clark-etal-2020-electra}. The Transformer model \cite{vaswani-etall-transformer} is in most cases at the core of these improvements. The self-attention mechanism of Transformer enables a better understanding of underlying dependencies in sentences.

However, the aforementioned techniques have not yet been tested on the RDF-to-text task. We therefore propose to combine the Transformer model with large pre-trainings to generate more accurate verbalization of the RDF triples. We chose to make use of external corpora during pre-training to reach a better generalization. We build two additional datasets for denoising pre-training and data augmentation purposes. This is motivated by the fact that participants' systems at previous WebNLG challenge suffered from high drops in performance when evaluated on new entities and predicates not encountered during the training phase \cite{gardent-etal-2017-webnlg}.  Thus, by incorporating external corpora, we expect our system to outperform systems solely trained on the WebNLG dataset. We evaluate our approach using BLEU \cite{papineni-etal-2002-bleu}, METEOR \cite{banerjee-lavie-2005-meteor}, chrf++ \cite{popovic-2015-chrf} and BLEURT \cite{sellam-etal-2020-bleurt}. As mentioned in \cite{post-2018-call}, BLEU is a parameterized metric that may exhibit wild variations. The use of multiple metrics gives better outlooks on the general performance.

The reminder of this paper is organized as follows. Section \ref{related_work_section} gives an overview of existing methods to translate RDFs into their corresponding textual content. After introducing the task and the WebNLG datasets in Section \ref{webnlg_challenge}, we introduce our approaches for the challenge in Section \ref{proposed_approach}. The different training phases are presented in Section  \ref{training_section}. We show our results in Section \ref{experiments_section}. Finally, we conclude in Section \ref{conclusion_section}.

\section{Related Work}
\label{related_work_section}
Following the traditional NLG pipeline \cite{reiter-nlg-pipeline-1997}, early works on factual data verbalization mainly focused on the creation of rules or templates to produce the textual output. \citet{sun-mellish-2006, sun-mellish-2007} have shown that most of the useful information for the generation is brought by the rich linguistic information of the RDF properties, \textit{i.e.} predicates. They have noticed that properties can be classified into 6 categories based on their pattern. For instance, some predicates may be a concatenation of a verb and a noun like \textit{hasEmail}, or starting with a verb and finishing with an adjective like \textit{hasTimeOpen}. They established handcrafted rules to construct linguistic forms for each category. This technique allows a domain independent verbalization since relying only on the pattern of predicates. On the contrary,  \citet{cimiano-etal-2013-exploiting} used domain-dependent ontology lexicon to proceed to a fine-grained and specific verbalization of mentioned concepts. Although some methods tried to automatically learn those templates \cite{duma-klein-2013-generating}, a major disadvantage of such approaches is the need of handcrafted rules and the poor generalization power. 

The main downside of KB verbalization is the lack of paired data. In spite of the impressive success of Deep Learning (DL) methods, few attempts could be made toward RDF-to-text task due to data scarcity. To address this issue, \citet{gardent-etal-2017-creating} came up with a new dataset which includes RDFs extracted from DBPedia\footnote{https://wiki.dbpedia.org/} with their corresponding translations in natural language. To further stimulate the development of KB verbalizer, \citet{gardent-etal-2017-webnlg} set up the WebNLG challenge 2017, welcoming new techniques for the RDF-to-text task. Among the proposed strategies, DL-based models outperformed rules-based systems \cite{gardent-etal-2017-webnlg}. Neural approaches relied almost on the attention-based encoder-decoder architecture \cite{sutskever-seq2eq-2014, bahdanau-attention}. The encoder-decoder framework showed promising results, with some limitations on unseen domains. 

Following WebNLG 2017, multiple works were conducted to cope with the witnessed drawbacks of participants' system. \citet{zhu-etal-inverse} argued that Kullback-Leibler (KL) divergence should be avoided. Minimizing the KL divergence contributes to more diversity in generated samples, at the cost of quality. As a matter of fact, diversity is unnecessary for KBs verbalization since it is constrained on specific entities and relations. Therefore, they proposed to optimize the inverse KL divergence to generate higher-quality textual description. Another flaw appears in the linearization of input triples. Such transformation gets rid of the intrinsic structure of the RDFs graph which may lead to inconsistency or relational mismatch in predicted sentences. To address this question and upon the rising popularity of Graph Neural Networks (GNNs), graph-based solutions were presented. \citet{trisedya-etal-2018-gtr} introduced a GTR-LSTM triple encoder. The idea is to traverse the graph in a specific manner such that vertices are better ordered to leverage the \textit{intra-} and \textit{inter-}triples relationships. 

More recently, Graph Neural Networks (GNNs) have raised a lot of attention to capture dependencies between nodes of a graph \cite{graph-neural-nets-survey}. The message passing scheme offers the ability to better model relations nodes and their neighborhoods. Well-suited for the generation from structured data, \citet{marcheggiani-perez-beltrachini-2018-deep} made use of a Graph Convolutional Network (GCN) to explicitly encode the input structure of the triples instead of a straightforward linearization. More precisely, they used an encoder-decoder architecture composed of a GCN encoder and a LSTM decoder. They further applied a copy mechanism from \cite{see-etal-2017-get} to copy words from the input RDF triples. The resulting performances confirmed the benefit of the graph-based encoder. Following this conclusion, several papers proposed graph-based encoder as a solution \cite{grah-augmented-ijcai2020, zhao-etal-2020-bridging, moussallem2020nabu}.

Meanwhile, in contrast to the conditional text generation paradigm, \citet{iso-etal-2020-fact} considered the KB verbalization as a text editing task with their \textsc{facteditor} model. Text editing consists in applying a sequence of transformations on an input text to produce a particular output. \textsc{facteditor}, a simple BiLSTM model, creates a verbalization candidate, so-called \textit{draft text}, from which the edition process starts. Three actions can be made on a candidate word: \textit{keeping} it, \textit{dropping} it or \textit{generating} a new one.

Recent studies also explore the fusion of both NLU and NLG into one unified model. The RDF-to-Text and Text-to-RDF can be seen as the inverse task from one another. Inspired by the advancement from computer vision \cite{zhu-etal-2017-unpair}, \citet{tseng-etal-2020-generative} and \citet{Guo-CycleGT-2020} provide a cycle training framework to learn both tasks simultaneously.

Nevertheless, none of the above techniques use large pre-training neither take advantage of the Transformer model. We propose in this paper to investigate the effect of massive pre-training coupled with the vanilla Transformer for the RDF-to-Text task.

\section{The WebNLG Challenge}
\label{webnlg_challenge}
In this section we present the WebNLG task and datasets.
\subsection{Task Definition}
\label{task_definition_section}
We call \emph{RDF verbalization} and \emph{RDF lexicalization} the mapping of the RDFs triples to their descriptive text. It is a variant of the machine translation task where the input is a list of triples instead of a raw text.
Let $T$ be a set of RDFs triples of size $N$, with $t_{i}$ the $i^{th}$ triple, $i \in [1 ... N]$. We write each triple $t_{i}$ as $\langle s_{i}\textit{; }p_{i}\textit{; }o_{i} \rangle$ with $s_{i}$, $p_{i}$ and $o_{i}$ the subject, predicate and object of the $i^{th}$ triple respectively. The $k^{th}$ verbalization of $T$ is denoted $v^{k}$. The RDF-to-text task consists in mapping the set $T$ to one of its possible lexicalization $v_{k}$. The generated verbalization should be grammatically and semantically correct. Regarding the knowledge, it should also contain the same level of information as the input triples.

\subsection{Datasets}
\label{webnlg_data_section}
Initially, \citet{ gardent-etal-2017-creating, gardent-etal-2017-webnlg} fashioned a corpus filled with \textit{triplesets}, \textit{i.e.} sets of RDF triples, and their possible lexicalizations, both in English. Each tripletset is related to a particular DBpedia category (\textit{e.g. Astronaut}) and can contain up to 7 triples.
For this new version of the WebNLG 2020, some changes were carried out to encourage participation to the competition. First, the unseen testing categories of the original WebNLG dataset are now part of the training set. The new extra category \textit{Company} is also included reaching now 16 domains for training. Moreover, to go beyond monolingualism, the second WebNLG challenge propose a bilingual corpus by inserting Russian translations for some RDF triples beside English ones. In our work, we however only focus on English verbalization. The support of multiple languages is part of our for future work.

The training set contains 13,211 unique triplesets for which different possible lexicalization can be given. The more triples in a tripleset, the higher the number of possible lexicalization. Totally, we can list out 35,426 RDFs-Text pairs. Note that a corresponding verbalization of a set of RDFs is not necessary a single sentence. Multiple sentences may encompass the RDFs information and co-references may therefore appear. The official WebNLG testing set is divided into three parts for fine-grained evaluation: seen categories, unseen entities, and unseen categories subsets. We give the evaluation performance on those subsets in Section \ref{results_section}.


\section{Proposed Approaches}
\label{proposed_approach}
We propose to use a Transformer model with data augmentation. 
\subsection{The Transformer Model}
\label{transformer_section}

The Transformer model introduced by \citet{vaswani-etall-transformer} is an encoder-decoder neural network architecture that relies on the self-attention mechanism. In an encoder-decoder architecture, the encoder maps the input sequence to a sequence of continuous representations from which the decoder generates an output sequence \cite{sutskever-seq2eq-2014}. 
The Transformer model makes use of stacked layers of multi-head self-attention in both the encoder and decoder.

Self-attention allows longer dependencies to be handled than the seq2seq model that relies on recurrent LSTM or GRU cells \cite{sutskever-seq2eq-2014, bahdanau-attention}.
The self-attention mechanism seems well-suited to handle subject, predicate, and object dependencies within the RDF itself but also dependencies between triples. 

As highlighted in the original paper \cite{vaswani-etall-transformer}, the individual attention heads from the multi-head attention layers learn to perform individually different tasks. In particular, they appear to learn the syntactic and semantic structure of the sentences independently. \citet{marecek-rosa-2019-balustrades} has shown that the self-attention layers embed implicitly syntactic parse trees. Similarly, we hope that our model benefits from our pre-training stage by incorporating such syntactic information, while learning intra and inter-triples dependencies in the fine-tuning stage.

\subsection{Data Augmentation}
\label{data_augmentation_section}
A major shortcoming of the WebNLG dataset remains its size. The lack of diversity impinges on the generalization of unseen categories. We report only 747 and 385 distinct subject entities and predicates, respectively. Thus, our intuition is to enlarge our RDF-to-text mappings to handle multiple relations between different type of entities. To do so, we proceed in two phases. First, we collect a large set of sentences from which, in our second stage, triples will be extracted. We hereinafter give details for each stage.

\subsubsection{Sentence Collection}
We make use of Wikipedia that offers a massive and free amount of data to be leveraged. The WebNLG dataset was constructed from DBPedia database which itself relies on Wikipedia. The substantial advantage of including new Wikipedia content lies in that new named entities can then be encountered from much more heterogeneous domains. This allows for a better generalization and knowledge integration. As observed in language models \cite{devlin-etal-2019-bert, radford2019language, zhang-etal-2019-ernie}, knowledge is somehow assimilated by the model throughout training. 

We gathered 13,614 Wikipedia pages containing about 103 million sentences totally\footnote{We used the \textit{20200401} Wikipedia dump $\sim$ 18GB}. To alleviate the training process, we filter sentences that may impede generation quality. We discard a sentence (1) if it does not start with a upper case letter and does not end with a period, (2) if its length is longer than 50 or (3) if it contains special characters. After this first selection, 57 million (unique) sentences remain. Henceforth, we coin this set of sentences WS1. 

\subsubsection{Triple Extraction}
Triples extraction is challenging and non-trivial, demanding a sharp understanding of linguistic structure. For simplicity (and time-saving), we take advantage of the Stanford Open Information Extraction (Stanford OpenIE) tool\footnote{Available at \url{https://nlp.stanford.edu/software/openie.html}} to extract triples from WS1 sentences. Stanford OpenIE package is schema-free. That means no preliminary definition of the possible predicates is required, as opposed to usual RDFs extractor. The raw text linking two entities will be retrieved as the predicate. Unfortunately, the returned triples may be incomplete, false or alike. For example, for the input sentence \textit{"Barack Obama was born in Hawaii."}, the returned triples by Stanford OpenIE are  $\langle \textit{Barack Obama; was; born} \rangle$ and $\langle \textit{Barack Obama; was born in; Hawaii} \rangle$. The first output triple is spurious besides expressing somehow the same insight than the second triple. As a result, a filtering step is crucial to reduce false triples. We fashion simple rules to limit redundancy by comparing each retrieved triple with each other. Let $t_{i}$ and $t_{j}$ be two retrieved triples, with $i \ne j$. We linearize each triple by concatenating words from the subject, predicate and object. The linearized version of $t_{i}$ and $t_{j}$ are named $t_{i}^{l}$ and $t_{j}^{l}$, respectively. To detect if $t_{i}^{l}$ and $t_{j}^{l}$ are equivalent, we first verify if either $t_{i}^{l} \in t_{j}^{l}$ or $t_{j}^{l} \in t_{i}^{l}$. We further compute the \emph{Levenshtein distance} \cite{levenshtein-distance} to consider minor variations between triples. Two triples with an edit distance smaller or equal to 2 are also believed to be similar. If triples similarity is confirmed, the longest triple is kept because sharing the more information with the input sentence. The higher the lexical coverage, the better. Indeed, the risk is to miss essential information from the sentence. If input triples are considerably incomplete, the model will suffer from hallucination, \textit{i.e.} generating content not present in the given input. 

However, when $t_{i}^{l}$ and $t_{j}^{l}$ are too different but semantically identical, more sophisticated conditions should be exploited. For this purpose, we make use of BLEU \cite{papineni-etal-2002-bleu}. Extracted triples are derived from the same sentence. Therefore, a \textit{n-gram} based metric such as BLEU is a good choice to check if two triples are alike. We assume triples to be analogous if their BLEU score is greater than 50. In such case, we keep the triple maximizing its BLEU score with the reference sentence for better coverage, as explained previously. The final collection of remaining triples-sentences pairs is called ST1.

Nevertheless, the detection of erroneous information conveyed by the triples remain hard to operate. We are aware that mistaken triples jeopardize directly the performance of our model. That is why we leverage these augmented data in a pre-training step, as detailed in Section \ref{pretraining_section}.




\section{Training}
\label{training_section}
We suggest in our work to thoroughly combine a pre-training stage before training our Transformer model on the WebNLG dataset. We describe our learning design in the following subsections.

\subsection{Pre-Training Objectives}
\label{pretraining_section}
From word embeddings to recently massive language models, NLP has undoubtedly benefited from large pre-training. Many downstream tasks could leverage those models by an effortless fine-tuning. Recent pre-trained models like BART \cite{ peters-etal-2019-tune} harness unlabeled data to boost model performance in a self-supervised setting. The denoising autoencoder objective from BART has shown a significant performance gain. A denoising objective aims to reconstruct a corrupted input. When fine-tuned on a text generation task, BART has confirmed to be highly effective. We therefore adopt a similar approach using our voluminous Wikipedia sentences, \textit{i.e.} WS1 dataset. \citet{peters-etal-2019-tune} use an arbitrary noising function to permute, delete, and mask words in input. Our transformations differ from BART in that we want our corrupted input to contain factual information, similarly to what our model will be exposed to when fine-tuned on our RDF-to-text task. With this in mind, we choose to keep words with specific Part-Of-Speech (POS) tags like nouns, verbs, adjectives and adverbs. Words with other tags are removed. Modal verbs (\textit{e.g.  should}) and passive forms are ignored as well. As an example, let the Wikipedia sentence \textit{``In 1860 few of the streets north of 42nd had been graded."}. After our noising transformations, we obtain the corrupted input \textit{``1860 few streets north 42nd graded."}. We observe that remaining words stand for the \textit{semantic mass} of the sentence from which we need to lexicalize and connect entailed concepts and entities properly.

At the difference of RDFs triples, the corrupted sentences have no inherent structure. In order to avoid too much divergence between pre-training and fine-tuning inputs' representation, we follow the denoising pre-training with another pre-training process on a RDF-to-text task. To do so, We use our constructed ST1 dataset which contains triple-based inputs, as detailed in Section \ref{data_augmentation_section}. We expect that these two pre-training steps will improve our ability to generate faithful and consistent verbalization for the WebNLG Challenge.

\subsection{Fine-tuning Settings}
\label{finetuning_section}


We coerce our model to use a WebNLG-based vocabulary during pre-training so that the same model can be straightly fine-tuned without any vocabulary discrepancy. Technically, we load the last checkpoint from pre-trained model and directly start to fine-tune it by reseting the 
optimizer and setting the new data loader to the WebNLG training set.  
We also tried without resetting the optimizer and obtained similar performances.

In addition, we apply a Curriculum Learning (CL) approach for faster convergence and better local minima through the optimization process.
CL was proposed by \citet{bengio-2009-curriculum}, motivated by the learning process of humans and animals. The core idea is to gradually increase the complexity of samples seen during training instead of a random selection. In our case, we define complexity as the number of RDF triples in the input, \textit{i.e.} $N$. The higher the number of triples, the higher the complexity. We sort the WebNLG pairs such that easier examples come first and then harder examples afterwards. Our curriculum approach differs in that both easier and harder examples are seen within the same epoch, but in a gradually order. In \cite{bengio-2009-curriculum}, more complex samples are progressively added during training. 
To witness the effect of increasing complexity while training, we fine-tune our pre-trained models with and without curriculum learning. 





\section{Experiments}
\label{experiments_section}

\begin{table*}[t]
\centering
\small

\begin{tabular}{c|l|l}
\hline
\textbf{Task} & \textbf{Input sequence} & \textbf{Output sequence} \\
\hline 
\multirow{4}{*}{WebNLG} &  \begin{tabular}[c]{@{}l@{}} \angles{subject} Italy \angles{predicate} capital \angles{object} Rome \angles{eot}  \end{tabular} & Rome is the capital of Italy . \\ 
\cline{2-3} 
& \begin{tabular}[c]{@{}l@{}} \angles{subject} Bionico \angles{predicate} course \angles{object} Dessert \\ \angles{eot} \angles{subject} Bionico \angles{predicate}  ingredient \angles{object} \\ Raisin \angles{eot}  \end{tabular}& Bionico is a dessert which contains raisins .  \\
\hline
\multirow{1}{*}{WS1} & \begin{tabular}[c]{@{}l@{}} He died he\bpe art f\bpe ail\bpe ure h\bpe   ospital Oc- \\ tober 5 2014 \end{tabular} & \begin{tabular}[c]{@{}l@{}}He died of he\bpe art f\bpe ail\bpe ure \\ at the h\bpe ospital on October 5 , 2014 . \end{tabular} \\ 
\hline
ST1 & \begin{tabular}[c]{@{}l@{}}  \angles{subject} He \angles{predicate} retired \angles{object} 199\bpe 0 \angles{eot} \end{tabular} & He retired in 199\bpe 0 . \\ 
\hline
\end{tabular}
\caption{\label{table:ex_data}
\textbf{Preprocessed samples} fed to the Transformer model during pre-training or fine-tuning. The triples are concatenated into one sequence. We used Moses tokenizer and a subwords encoding, the subwords are divided by \emph{\bpe}. Special tokens are used to keep the RDFs triplet format : \angles{subject}, \angles{predicate}, \angles{object} and \angles{eot}. In the case of the denoising task (\textbf{WS1}) we do not have these special tokens. We left the data trucased and no delexicalization is applied.
}
\end{table*}

We evaluate our models on the WebNLG Dataset test set. We compare 8 Transformers models corresponding to 8 different training settings.

\subsection{Training and decoding settings}
In the following, we present the preprocessing, the training phase and the decoding pipeline.

\paragraph{Preprocessing}
For preprocessing, we used the Moses tokenizer\footnote{Available on github \url{https://github.com/moses-smt/mosesdecoder}} and subword segmentation following \newcite{sennrich-etal-2016-neural} with the subword-nmt 
library\footnote{Available on github \url{https://github.com/rsennrich/subword-nmt}}. We left both the input triples and output text true-cased.


In the context of the WebNLG challenge, we considered a transduction strategy \cite{gammerman-transduction}. We restricted the vocabulary to the WebNLG dataset (training, validation and testing set). This learning scheme aims at performing well on a specific set, and not necessary to generalize.

The training process remained the same for all our experiments. Only the data loaded in the batches during training were changed.

To deal with the RDF triples format, we added to the vocabulary four special tokens, namely \angles{object}, \angles{subject}, \angles{predicate} and \angles{eot} (end of triple), that we used as separators within and between triples. In the case of multiple triples, we built the Transformer input sequence by concatenating triples one after the other. We used this input format on both ST1 and WebNLG.

For the denoising pre-training, we used the WS1 dataset. The transformed sentences (see Section \ref{pretraining_section}) were directly fed to the Transformer. The model thus has to reconstruct the incomplete sentences.

Examples of the preprocessed samples for the different training settings are reported in Table \ref{table:ex_data}.

\paragraph{Training}

We used the Transformer implementation from \textsc{fairseq} library \cite{ott2019fairseq}\footnote{We used the compiled version 0.9.0 from \url{https://github.com/pytorch/fairseq}} with the \emph{transformer\_base} hyper-parameters set from \cite{vaswani-etall-transformer}. We optimized the weights of our neural networks using an \textsc{Adam} optimizer and a label smoothed cross entropy loss.

We made 10 epochs of pre-training and stopped fine-tuning when the performance with the BLEU score on the validation set did not improved after 30 epochs. At the end, we kept the model that achieved the best \textsc{BLEU} on the validation set.

To study the impact of curriculum learning, we launched fine-tuning with and without it. In the former case, we prevented the shuffling of the batches for 30 epochs. The data being sorted by number of triples, the model had to deal first with simple samples, and then with more complex ones as the learning did progress. In the results, we report this setting as CL (for Curriculum Learning).


\paragraph{Decoding}
For decoding, we did a beam search with a beam of size 5. We merged the subwords back into words and detokenized. 

\subsection{Evaluation setting}
For evaluation, we used the official WebNLG evaluation script\footnote{Available on github \url{https://github.com/WebNLG/GenerationEval}}. The metrics we used to compare our models are BLEU, METEOR, chrf++ and BLEURT. BLEURT metric has been recently proposed and proven to be well-correlated with human judgments. Relying on BERT's contextual embeddings, BLEURT offers semantically robust feedback. The \emph{n-gram}-based evaluation techniques such as BLEU, METEOR or chrf++ are additional metrics to judge the generation quality. When used together, they give good assessment of the generation quality of our system.

Unlike in 2017, this time the generated sentences must be detokenized and true-cased. Our models are therefore not directly comparable with the models of the previous version of the challenge. 

Also, as the test base is not available, we do not know how many references our generated sentences are compared to.

\subsection{Results}
\label{results_section}

\begin{table*}[hbt!]
\centering
\begin{tabular}{ccc|c||cccc}
\hline
\multicolumn{8}{c}{\textbf{Seen categories}} \\
\hline
{WebNLG} & {WS1} & {ST1} & {CL} & {BLEU} & {METEOR} & {chrf++} & {BLEURT} \\ \hline
\checkmark &&&& 55.24 & 0.401 & 0.680 & 0.56 \\
\checkmark &\checkmark&&& 57.3 & 0.417 & 0.701 & 0.58 \\
\checkmark & & \checkmark && \textbf{59.32} & \textbf{0.428} & \underline{0.712} & \underline{0.6} \\
\checkmark & \checkmark & \checkmark && 57.49 & 0.420 & 0.702 & 0.59 \\
\hline
\checkmark &&& \checkmark & 54.78 & 0.399 & 0.676 & 0.56 \\
\checkmark &\checkmark && \checkmark & 56.94 & 0.417 & 0.701 & 0.58 \\
\checkmark &&\checkmark& \checkmark & 58.36 & 0.422 & 0.703 & \textbf{0.61} \\
\checkmark &\checkmark &\checkmark& \checkmark & \underline{58.81} & \underline{0.427} & \textbf{0.713} & \underline{0.6} \\
\hline
\multicolumn{8}{c}{} \\
\hline
\multicolumn{8}{c}{\textbf{Unseen entities}} \\
\hline
{WebNLG} & {WS1} & {ST1} & {CL} & {BLEU} & {METEOR} & {chrf++} & {BLEURT} \\ 
\hline
\checkmark &&&& 12.9 & 0.167 & 0.319 & -0.62 \\
\checkmark &\checkmark&&& 29.16 & 0.301 & 0.518 & 0.1 \\
\checkmark & & \checkmark && \textbf{35.77} & \textbf{0.326} & \textbf{0.565} & \textbf{0.26} \\
\checkmark & \checkmark & \checkmark && 32.33 & 0.310 & 0.535 & 0.18 \\
\hline
\checkmark &&& \checkmark & 11.94 & 0.156 & 0.295 & -0.68 \\
\checkmark &\checkmark && \checkmark & 28.83 & 0.295 & 0.509 & 0.05 \\
\checkmark &&\checkmark& \checkmark & 31.99 & 0.312 & 0.537 & 0.16 \\
\checkmark &\checkmark &\checkmark& \checkmark & \underline{32.69} & \underline{0.315} & \underline{0.542} & \underline{0.19} \\
\hline 
\multicolumn{8}{c}{} \\
\hline
\multicolumn{8}{c}{\textbf{Unseen categories}} \\
\hline
{WebNLG} & {WS1} & {ST1} & {CL} & {BLEU} & {METEOR} & {chrf++} & {BLEURT} \\ 
\hline
\checkmark &&&& 11.17 & 0.162 & 0.310 & -0.64 \\ 
\checkmark &\checkmark&&& 21.02 & 0.265 & 0.452 & -0.06 \\ 
\checkmark & & \checkmark && \underline{23.26} & \textbf{0.288} & \textbf{0.485} & \textbf{0.03} \\
\checkmark & \checkmark & \checkmark && \textbf{23.42} & 0.275 & \underline{0.469} & \underline{-0.02}\\
\hline
\checkmark &&& \checkmark & 12.45 & 0.156 & 0.298 & -0.63 \\
\checkmark &\checkmark && \checkmark & 21.84 & 0.263 & 0.451 & -0.04 \\ 
\checkmark &&\checkmark& \checkmark & 22.84 & 0.273 & 0.465 & -0.05 \\
\checkmark &\checkmark &\checkmark& \checkmark & 22.72 & \underline{0.276} & 0.466 & \underline{-0.02} \\
\hline 
\end{tabular}

\caption{\label{results_table}
Ablation study: Automatic results on the official WebNLG test set for different learning strategy. For each experiment, we provide testing performance on seen categories (top), unseen entities (middle) and unseen categories (bottom). Bold and underlined values correspond to the best and second-best results respectively. 
}
\end{table*}

To gauge the influence of our different learning approaches, we conducted an ablation study. We defined our baseline as the Transformer solely trained on WebNLG dataset, without curriculum. We want to assess the influence of pre-training, data augmentation and curriculum learning compared to the baseline. To provide a fair and detailed analysis, we evaluated models on seen and unseen domains to shed light on the models' generalization ability. Results are given in Table \ref{results_table}.

In Table \ref{results_table}, when comparing our baseline with pre-training strategies for each category (no curriculum), we note an average rise of 3.07, 19.6 and 10.97 respectively in BLEU when pre-training. Similar variations can be noted with METEOR, chrf++ and BLEURT metrics, albeit less striking. Based on BERT's contextual embeddings, BLEURT gives a good estimate of the semantic correlation between prediction and references. Most of the time, \emph{n-gram}-based and semantic metrics show perfect harmony. Top values of BLEURT are obtained for the same models than \emph{n-gram}-based metrics. Therefore, all metrics tend to correlate proving a good agreement. The high improvements over unseen domains are easily explained due to the diversity of augmented data. New entities and domain-specific lexicon encountered better help to model out-of-distribution data relations. Thus, such results underline usefulness of external corpora and strengthen the need of pre-trained model to lexicalize KBs.

On seen categories, our baseline give an acceptable BLEU score of 55.24. However, for out-of-domain generation, all models demonstrate severe shortcomings. Tested on unseen entities, our baseline shows a BLEU drop of 42.34 to reach 12.9 BLEU. We witness similar and even more important loss in unseen categories. In the case that predicates are unknown to the model, it is hard to generate consistent description of the input RDFs. From seen categories to unseen categories, our baseline is nearly penalized by a factor of 5. Such effect is tempered with our pre-training approaches. The average drop in BLEU of pre-trained models (without curriculum learning) is 25.6 from seen categories to unseen entities, and 35.5 from seen to unseen categories.
 
Unexpectedly, when a curriculum learning approach is used, we witness drops in performance. This is counter-intuitive and opposed to previous experiment on our validation set. CL seems to help lightly when the model was pre-trained with both external corpora WS1 and ST1. We propose to investigate the reason of such outcome in future work.

Interestingly, best results are revealed with a pre-training on ST1, exclusively. With 5 times less data, ST1 leads to better performance. The extracted triples surely include inaccurate triples. Notwithstanding the imperfect quality of the ST1 dataset, its use contributes to generalization ability. 
On the contrary, we report that denoising pre-training does not show satisfying results when combined with our pre-training on the ST1 dataset, leading eventually to negative effect. This may be caused by an input distribution discrepancy between WS1 and ST1 input. Denoising pre-training doesn't require triples as input but a noisy sentence. Mismatch between this representation and the triples linearization may be the culprit of such side-effect. 

As for the WebNLG 2020 challenge, participants were requested to submit a single model for evaluation. In our case, we decided to submit our Transformer pre-trained with both WS1 and ST1 (without curriculum). After an analysis of the generated text, the model trained with much more data has a tendency to be much more fluent and aggregate information better. \citet{castro-ferreira20:_2020_bilin_bidir_webnl} report human evaluation based on different criteria: data coverage, relevance, correctness, text structure and fluency. For each criterion, a value between 0 (complete disagreement) and 100 (complete agreement) is given. The scores are normalized and then clustered into groups such that models of a same cluster do not show any significant statistical differences in their scores\footnote{Details of the evaluation procedure is outlined in \cite{castro-ferreira20:_2020_bilin_bidir_webnl} and scores are publicly available at \url{https://gerbil-nlg.dice-research.org/gerbil/webnlg2020resultshumaneval}}. When tested on seen categories, we note that our submitted model is competitive with other teams' models. However, on unseen data, although a significant improvement over a simple training of a Transformer, our model shows limitations compared to other participants. A lower rank is systematically observed for both unseen entities and unseen categories.
We assume that a delexicalization step and a much massive pre-training as \cite{devlin-etal-2019-bert, radford2019language} may help to improve our generalization ability on unseen domains. This direction is also to be explored in future work.

\section{Conclusion}
\label{conclusion_section}
We proposed to use the Transformer model with data augmentation for the RDF-to-text task on the WebNLG dataset. We took advatange of Wikipedia to build new datasets for different pre-training strategies. We investigated the effect of denoising autoencoder objective and data augmentation as pre-training approaches. We studied the impact of curriculum learning while fine-tuning our model on WebNLG data. Further analysis demonstrated that pre-training is highly beneficial for knowledge verbalization. Although noisy, our massive data augmentation contributed to generate much accurate textual content. As future work, we will perform a deep analysis of the robustness of our models to unseen entities and categories.
\section{Acknowledgment}
This work was partially funded by the ANR Cifre conventions N°2020/0400 and N°2018/1559 and Orange-Labs Research. This work was granted access to the HPC resources of IDRIS under the allocation \textit{2020-AD01101126} made by GENCI. We also thank Thibault Cordier and Imen Akermi for the fruitful discussions.

\bibliography{anthology,acl2020}
\bibliographystyle{acl_natbib}

\appendix

\end{document}